\documentclass{INTERSPEECH2023}

\interspeechcameraready

\usepackage{cite}
\usepackage{tikz}
\usepackage{pgfplots}
\pgfplotsset{compat=1.3}
\usepackage{multirow}
\usepackage{xcolor}
\usepackage{subcaption}

\newcommand{\norm}[1]{\lVert #1 \rVert} % norm: double vertical bars

\newcommand{\Lc}{\mathcal{L}}

\newcommand{\Sc}{\mathcal{S}}

\newcommand{\Uc}{\mathcal{U}}

\newcommand{\Eb}{\mathbb{E}}

\newcommand{\Rb}{\mathbb{R}}

\newcommand{\uv}{\mathbf{u}}
\newcommand{\vv}{\mathbf{v}}

\newcommand{\xv}{\mathbf{x}}

\newcommand{\zv}{\mathbf{z}}

\newcommand{\Wv}{\mathbf{W}}
\newcommand{\Xv}{\mathbf{X}}

\newcommand{\alphav     }{\boldsymbol \alpha     }

\newcommand{\thetav     }{\boldsymbol \theta     }

\newcommand{\lambdav    }{\boldsymbol \lambda    }

\title{DPHuBERT: Joint Distillation and Pruning of Self-Supervised Speech Models}
\name{Yifan Peng$^1$, Yui Sudo$^2$, Shakeel Muhammad$^2$, Shinji Watanabe$^1$}
\address{
  $^1$Carnegie Mellon University, Pittsburgh, PA, USA\\
  $^2$Honda Research Institute Japan Co., Ltd., Saitama, Japan}
\email{yifanpen@andrew.cmu.edu, \{yui.sudo, shakeel.muhammad\}@jp.honda-ri.com, shinjiw@ieee.org}

\begin{document}

\maketitle

\begin{abstract}
Self-supervised learning (SSL) has achieved notable success in many speech processing tasks, but the large model size and heavy computational cost hinder the deployment. Knowledge distillation trains a small student model to mimic the behavior of a large teacher model. However, the student architecture usually needs to be manually designed and will remain fixed during training, which requires prior knowledge and can lead to suboptimal performance. Inspired by recent success of task-specific structured pruning, we propose DPHuBERT, a novel task-agnostic compression method for speech SSL based on joint distillation and pruning. Experiments on SUPERB show that DPHuBERT outperforms pure distillation methods in almost all tasks. Moreover, DPHuBERT requires little training time and performs well with limited training data, making it suitable for resource-constrained applications. Our method can also be applied to various speech SSL models. Our code and models will be publicly available.
\end{abstract}
\noindent\textbf{Index Terms}: model compression, knowledge distillation, structured pruning, self-supervised learning

\section{Introduction}
\label{sec:intro}

Self-supervised speech representation learning (speech SSL) has achieved remarkable results in various tasks~\cite{wav2vec2, hubert, xlsr, wavlm, baevski2021unsupervised, superb, ssl-review, ssl-for-asr, ssl-for-se-ss, ssl-for-slu}. However, speech SSL models are usually large and slow, making them unsuitable for real-world applications with limited resources. Compressing speech SSL has become an important topic. A popular method is knowledge distillation~\cite{hinton2015distilling}, which trains a small student model to match the outputs of a large teacher model. Prior studies such as DistilHuBERT~\cite{distilhubert} and FitHuBERT~\cite{fithubert} have achieved promising results with different student models. Another work~\cite{deep-vs-wide} shows that the student architecture affects its performance substantially, even if the model size is similar. However, in distillation methods, the student architecture is pre-specified and remains unchanged, which needs special expertise and might lead to suboptimal results. In contrast, pruning~\cite{pruning-survey, liu2018rethinking} automatically discovers a compact subnetwork from a large model, which has been explored in natural language processing (NLP)~\cite{head-pruning, Fan2020Reducing, asapp-pruning, cofi} and speech processing~\cite{structured-prune-rnn,prune-lstm,deliang-compressing-enhancement, parp, yifan-pruning}. Previous pruning methods for speech SSL focus on specific downstream tasks such as automatic speech recognition (ASR)~\cite{parp, yifan-pruning} and spoken language understanding (SLU)~\cite{yifan-pruning}. It is unclear how they will perform in task-agnostic compression, which is more challenging because the model needs to capture various aspects of speech including content, speaker, semantics and paralinguistics~\cite{superb}.

In this work, we propose DPHuBERT, a task-agnostic compression method based on joint \textbf{D}istillation and \textbf{P}runing. It allows the student architecture to be learned during distillation. Experiments on SUPERB~\cite{superb} show that our method outperforms prior distillation methods in almost all tasks. Our method also performs well for various speech SSL models such as HuBERT Base~\cite{hubert}, WavLM Base+~\cite{wavlm} and HuBERT Large~\cite{hubert}, even with limited training resources. We will submit our results to the SUPERB leaderboard and release the code and models publicly for reproducibility: \url{https://github.com/pyf98/DPHuBERT}.

\begin{figure*}[t!]
     \centering
     \hfill
     \begin{subfigure}[b]{0.495\linewidth}
         \centering
         \includegraphics[scale=0.42]{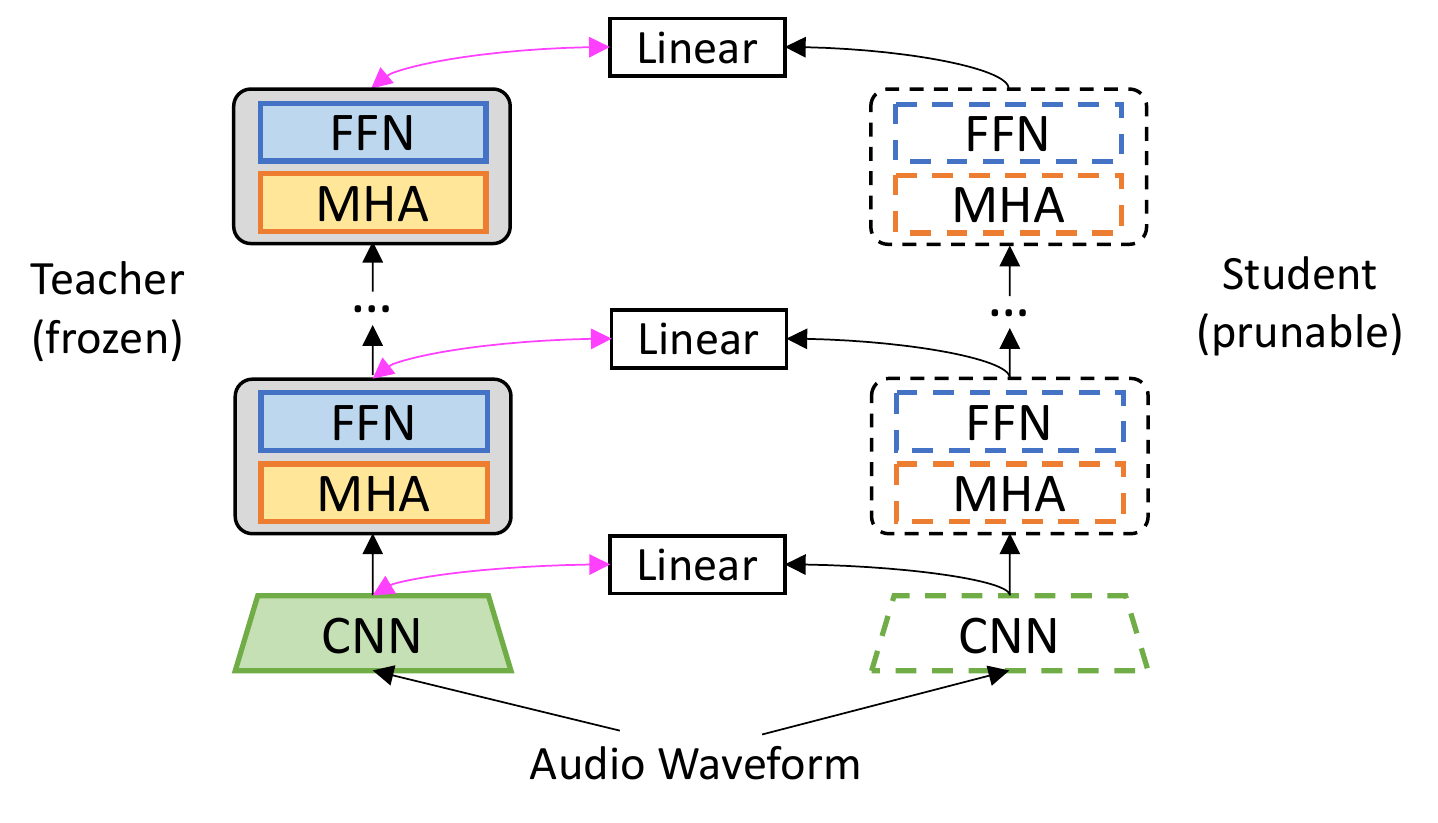}
         \vskip -0.12in
         \caption{Step 1: jointly distill and prune the student model.}
         \label{fig:training-step1}
     \end{subfigure}
     \hfill
     \begin{subfigure}[b]{0.495\linewidth}
         \centering
         \includegraphics[scale=0.42]{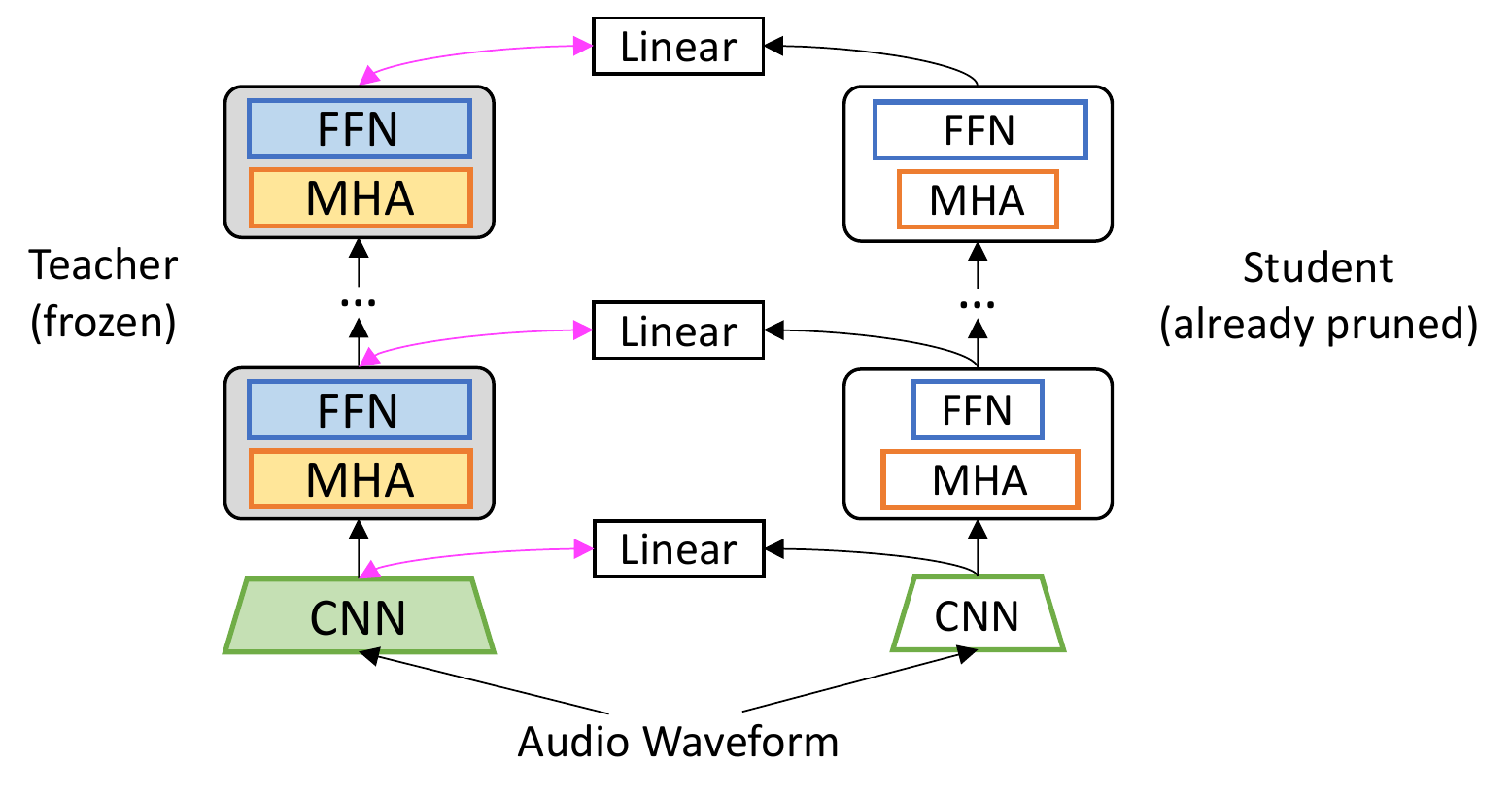}
         \vskip -0.12in
         \caption{Step 2: further distill the already pruned model.}
         \label{fig:training-step2}
     \end{subfigure}
     \hfill
    \vskip -0.12in
    \caption{Two training steps of our task-agnostic compression method, DPHuBERT. (a) The student model is initialized from the teacher model and is jointly distilled and pruned to produce a smaller model which meets a pre-specified sparsity ratio. (b) The already pruned student model is further distilled for better performance. To obtain DPHuBERT (24M) from HuBERT Base (95M), the two steps take around 18 and 6 GPU hours, respectively. (Dashed modules are prunable, i.e., their architectures can evolve during training.)}
    \label{fig:training-steps}
    \vskip -0.2in
\end{figure*}

\section{Background and related work}

\subsection{Architectures of speech SSL}

Speech SSL models such as wav2vec 2.0~\cite{wav2vec2}, HuBERT~\cite{hubert} and WavLM~\cite{wavlm} share a similar architecture. The model consists of a convolutional feature extractor (CNN) and a Transformer~\cite{transformer} encoder. The CNN has seven temporal convolutions with normalizations and activations. The Transformer encoder contains 12 layers with hidden size 768 for base models and 24 layers with hidden size 1024 for large models. Each layer is composed of a multi-head self-attention (MHA) and a position-wise feed-forward network (FFN).

\subsection{Compression methods for speech SSL}

\label{subsec:background-distill}

\noindent\textbf{Distillation.}
Distillation methods optimize a small student model to match the targets generated by a large teacher model. Since different layers of speech SSL capture different information~\cite{ankita-analysis-ssl}, the student model needs to learn both final and intermediate representations of the teacher~\cite{distilhubert, fithubert, deep-vs-wide}. DistilHuBERT~\cite{distilhubert} trains a shallow student model by mapping the last student layer to multiple intermediate teacher layers. FitHuBERT~\cite{fithubert} learns a deep and thin student model through layer-to-layer mapping. Another work~\cite{deep-vs-wide} compares prediction-layer and layer-to-layer distillation using various student architectures. It shows that the architecture of a student model affects its performance, even when the model size is kept similar. It also finds that deeper networks perform better with layer-to-layer distillation, likely because it explicitly aligns intermediate layers. These observations have inspired our work which allows the student architecture to evolve during distillation.

\noindent\textbf{Pruning.}
Pruning methods identify and remove redundant parameters from a pre-trained model. Unstructured pruning removes individual parameters (e.g., a connection between neurons) by setting them to zero, which requires sparse matrix computation libraries to achieve actual speedup, whereas structured pruning removes groups of parameters (e.g., an attention head or even an entire layer), which directly reduces the model size and computational cost.
For speech SSL, PARP~\cite{parp} is a magnitude-based unstructured pruning method which prunes the Transformer encoder. It improves downstream tasks like low-resource ASR. HJ-Pruning~\cite{yifan-pruning} is a structured pruning method that jointly prunes heterogeneous components (i.e., CNN and Transformer) of speech SSL models. It significantly reduces the total computation while retaining good performance in ASR and SLU. These methods deal with specific downstream tasks, but do not investigate the universal speech representations. Our work focuses on task-agnostic structured pruning of speech SSL. As there is no labeled data for normal supervised training, we employ a distillation objective along with pruning.

\noindent\textbf{Once-for-all training.}
Compression methods typically generate a single model with a pre-determined size. LightHuBERT~\cite{lighthubert} deploys once-for-all training~\cite{Once-for-All} to obtain numerous weight-sharing subnets, which shows very strong performance on SUPERB~\cite{superb}. However, it requires an expensive two-stage training process and an advanced distillation loss inspired by data2vec~\cite{data2vec}. According to the authors, compressing HuBERT Base already takes 2k GPUs hours (i.e., 62 hours with 32 V100 GPUs and 19 hours with 8 GPUs for two stages, respectively), which is prohibitive for academic researchers and small businesses. Unlike LightHuBERT, our work aims to compress an existing speech SSL model to a specific sparsity ratio within a manageable amount of training time, which is consistent with the standard setup of prior distillation methods~\cite{distilhubert, fithubert}.

\section{DPHuBERT}

\subsection{Training procedure}
\label{subsec:training-procedure}

Figure~\ref{fig:training-steps} illustrates our training procedure consisting of two steps. In Step 1, the student model is initialized from the teacher and is jointly distilled and pruned to generate a smaller model with a pre-specified size. In Step 2, the already pruned student model is further distilled to improve performance. In both steps, only unlabeled speech data are utilized and the teacher is frozen.

\subsection{Distillation loss}
\label{subsec:distill-loss}

Unlike DistilHuBERT~\cite{distilhubert}, we use layer-to-layer distillation since the student initially has the same depth as the teacher (see Section~\ref{subsec:background-distill} for discussions).
Suppose the teacher has $N^{\text{tea}}$ Transformer layers with hidden size $d^{\text{tea}}$ and the student has $N^{\text{stu}}$ layers with hidden size $d^{\text{stu}}$. Let $\Xv_i^{\text{tea}}$ (shape $T\times d^{\text{tea}}$) and $\Xv_i^{\text{stu}}$ (shape $T\times d^{\text{stu}}$) be the output sequences of the $i$-th Transformer layers from the teacher and student, respectively, where $T$ is the sequence length. The distillation loss is:
\begin{align}
    \Lc^{\text{dis}} = \sum_{i \in \Sc} \Lc\left( \Xv_i^{\text{tea}}, \Xv_i^{\text{stu}} \Wv_i \right),
    \label{eq:distill-loss}
\end{align}
where $\Sc$ is a set of layers to match between the teacher and student models after a linear projection $\Wv_i$. We use $\Sc = \{0, 4, 8, 12\}$ for base models and $\{0, 8, 16, 24\}$ for large models. The $0$th layer is the output of CNN, which is also the input to the first Transformer layer. The loss function $\Lc$ measures the difference between two feature sequences, which can be $L_1$, $L_2$ or cosine distances~\cite{distilhubert, fithubert, deep-vs-wide}. We follow~\cite{distilhubert, deep-vs-wide} to use a combination of $L_1$ and cosine distances with equal weights.

% Table
\begingroup
\setlength{\tabcolsep}{4pt}

\begin{table*}[t!]
  \caption{Comparison of our method versus previous distillation methods on SUPERB~\cite{superb}. Our DPHuBERT and DPWavLM are compressed from publicly available HuBERT Base~\cite{hubert} and WavLM Base+~\cite{wavlm} checkpoints, respectively.}
  \label{tab:main-results}
  \vskip -0.13in
  \centering
  \resizebox {0.84\linewidth} {!} {
  \begin{tabular}{lccccccccccc}
    \toprule
    \multirow{2}{*}{Method} & \#Params & KS & IC & PR & ASR w/o LM & ER & QbE & SF & SID & ASV & SD \\ \cmidrule{2-12}
    & Millions & Acc$\uparrow$ & Acc$\uparrow$ & PER$\downarrow$ & WER$\downarrow$ & Acc$\uparrow$ & MTWV$\uparrow$ & F1$\uparrow$ / CER$\downarrow$ & Acc$\uparrow$ & EER$\downarrow$ & DER$\downarrow$ \\
    \midrule
    FBANK & 0 & 41.38 & 9.65 & 82.01 & 23.18 & 48.24 & 0.0058 & 69.64 / 52.94 & 20.06 & 9.56 & 10.05 \\
    wav2vec 2.0 Base~\cite{wav2vec2} & 95.04 & 96.23 & 92.35 & 5.74 & 6.43 & 63.43 &  0.0233 & 88.30 / 24.77 & 75.18 & 6.02 & 6.08 \\
    HuBERT Base~\cite{hubert} & 94.68 & 96.30 & 98.34 & 5.41 & 6.42 & 64.92 & 0.0736 & 88.53 / 25.20 & 81.42 & 5.11 & 5.88 \\
    WavLM Base+~\cite{wavlm} & 94.70 & 97.37 & 99.00 & 3.92 & 5.59 & 68.65 & 0.0988 & 90.58 / 21.20 & 89.42 & 4.07 & 3.50 \\
    \midrule
    \multicolumn{12}{l}{\textit{Compressed models using LibriSpeech 960h}} \\
    \midrule
    DistilHuBERT~\cite{distilhubert} & 23.49 & 95.98 & 94.99 & 16.27 & 13.37 & 63.02 & 0.0511 & 82.57 / 35.59 & 73.54 & 8.55 & 6.19 \\
    FitHuBERT~\cite{fithubert} & 22.49 & 96.27 & 91.25 & 13.32 & 12.09 & 59.82 & 0.0489 & 84.06 / 32.46 & 55.71 & 8.00 & 6.84 \\
    FitW2V2~\cite{fithubert} & 31.63 & 96.04 & 93.38 & 12.22 & 11.44 & 62.35 & 0.0475 & 86.65 / 29.40 & 64.71 & 6.65 & 6.44 \\
    12-Layer Half~\cite{deep-vs-wide} & 26.87 & \textbf{97.24} & 96.97 & 10.67 & 10.96 & 63.24 & 0.0604 & 86.11 / 30.93 & 69.52 & 6.13 & 6.81 \\
    3-Layer One~\cite{deep-vs-wide} & 30.58 & 96.69 & 94.15 & 13.34 & 12.23 & 63.95 & 0.0489 & 82.89 / 34.65 & 75.71 & 6.48 & 6.56 \\
    \textbf{DPHuBERT (ours)} & 23.59 & 96.36 & 97.92 & 9.67 & 10.47 & 63.16 & 0.0693 & 86.86 / 28.26 & 76.83 & \textbf{5.84} & 5.92 \\
    \textbf{DPWavLM (ours)} & 23.59 & 96.27 & \textbf{98.58} & \textbf{8.22} & \textbf{10.19} & \textbf{65.24} & \textbf{0.0874} & \textbf{87.68} / \textbf{26.11} & \textbf{82.11} & 5.98 & \textbf{5.53} \\
    \midrule
    \multicolumn{12}{l}{\textit{Compressed models using LibriSpeech 100h}}\\
    \midrule
    DistilHuBERT~\cite{distilhubert} & 23.49 & - & 93.17 & - & 14.77 & - & - & - & 69.46 & - & -\\
    FitHuBERT~\cite{fithubert} & 22.49 & 96.23 & 94.20 & 14.05 & 12.66 & 61.67 & 0.0579 & 83.41 / 34.00 & 54.24 & 7.88 & 7.19\\
    FitW2V2~\cite{fithubert} & 22.49 & 94.68 & 90.03 & 16.50 & 14.77 & \textbf{62.87} & 0.0380 & 81.95 / 34.74 & 51.65 & 7.43 & 6.94\\
    \textbf{DPHuBERT (ours)} & 23.57 & \textbf{96.36} & \textbf{97.42} & \textbf{10.02} & \textbf{11.38} & 62.78 & \textbf{0.0634} & \textbf{84.83} / \textbf{33.03} & \textbf{73.37} & \textbf{6.25} & \textbf{6.03}\\
    \bottomrule
  \end{tabular}
  }
  \vskip -0.22in
\end{table*}

\endgroup

\subsection{Joint distillation and structured pruning}
\label{subsec:joint-distill-prune}

Structured pruning of the student model is formulated as learning a sparse model through $L_0$ regularization~\cite{l0sparse-iclr18}, which has been explored in NLP~\cite{asapp-pruning, cofi} and speech~\cite{yifan-pruning}. The method will be briefly introduced below. For more comprehensive derivations, please refer to prior research~\cite{l0sparse-iclr18, asapp-pruning, cofi, yifan-pruning}.
Consider a frozen teacher model $f^{\text{tea}}(\cdot)$ and a student model $f^{\text{stu}}(\cdot; \thetav)$ with learnable parameters $\thetav = \{\theta_j\}_{j=1}^n$. Each $\theta_j$ is a group of prunable parameters (including convolution channels, attention heads, and FFN intermediate units) and there are $n$ groups in total. We define a binary variable $z_j$ (called mask) for each $\theta_j$. The masks $\zv$ follow a probability distribution $q(\zv;\alphav)$ with parameters $\alphav$. The regularized distillation objective is:
\begin{align}
\label{eq:loss-l0-reg}
    \min_{\thetav, \alphav} ~
    \Eb_{\zv \sim q}\left[\frac{1}{D} \sum_{k=1}^D \Lc^{\text{dis}}\left( f^{\text{tea}}(\xv_k), f^{\text{stu}}(\xv_k;\tilde{\thetav}) \right) + \lambda \norm{\tilde{\thetav}}_0 \right],
\end{align}
where $ \tilde{\thetav} = \{\tilde{\theta}_j\}_{j=1}^n$ and each $\tilde{\theta}_j = \theta_j z_j$. The unlabeled dataset with $D$ samples is $\{\xv_k\}_{k=1}^D$. $\lambda > 0$ is the regularization weight.
It is intractable to solve Eq.~\eqref{eq:loss-l0-reg} using gradient descent due to the discrete nature of masks $\zv$. To make the loss differentiable, Louizos et~al. propose a reparameterization trick which samples $\zv$ from the Hard Concrete distribution~\cite{l0sparse-iclr18}:
\begin{align}
\label{eq:hard-concrete}
\begin{split}
    & \uv \sim \Uc(0,1), 
    ~~\vv(\alphav) = \text{sigmoid} \left( \left( \log\frac{\uv}{1-\uv} + \log\alphav \right) / \beta \right), \\
    & \bar{\vv}(\alphav) = (r-l)\cdot\vv(\alphav) + l, 
    ~~\zv = \min(1,\max(0,\bar{\vv}(\alphav))),
\end{split}
\end{align}
where $\uv$ follows a uniform distribution in $[0,1]$. $\beta$ is a constant. $l < 0$ and $r > 0$ are two constants to stretch $\vv$ to $[l, r]$, and it is further clamped to $[0,1]$. Only $\alphav = \{\alpha_j\}_{j=1}^n$ are learnable parameters in this distribution.
With this trick, the objective in Eq.~\eqref{eq:loss-l0-reg} is differentiable and the regularization term has a closed-form expression~\cite{l0sparse-iclr18}:
\begin{align}
    \label{eq:expected-norm}
    \Eb_{\zv \sim q}\left[\norm{\tilde{\thetav}}_0\right]
    = \sum_{j=1}^n \text{sigmoid} \left( \log \alpha_j - \beta \log\frac{-l}{r} \right),
\end{align}
which represents the (expected) model size as a differentiable function of current parameters $\alphav$.

Now Eq.~\eqref{eq:loss-l0-reg} can be solved to learn a sparse subnet, but the final sparsity cannot be precisely controlled~\cite{asapp-pruning, cofi}. To explicitly control the final model size, prior studies~\cite{asapp-pruning, cofi, yifan-pruning} rewrite the optimization problem with an equality constraint:
\begin{align}
\begin{split}
    & \min_{\thetav, \alphav} ~
    \Eb_{\zv \sim q} \left[\frac{1}{D} \sum_{k=1}^D \Lc^{\text{dis}}\left( f^{\text{tea}}(\xv_k), f^{\text{stu}}(\xv_k;\tilde{\thetav}) \right) \right]\\
    & \text{ s.t. }~ s(\alphav) = t,
\end{split}
\label{eq:eq-constraint}
\end{align}
where $s(\alphav)$ is the current sparsity (percentage of pruned parameters) of the student model and $t$ is a pre-specified target sparsity. Note that $s(\alphav)$ can be computed based on Eq.~\eqref{eq:expected-norm} because the $L_0$ norm counts the remaining parameters. The optimization objective in Eq.~\eqref{eq:eq-constraint} can be further converted to a minimax problem using augmented Lagrangian~\cite{asapp-pruning}:
\begin{align}
\begin{split}
    \max_{\lambda_1, \lambda_2} \min_{\thetav, \alphav} ~
    & \Eb_{\zv \sim q} \left[\frac{1}{D} \sum_{k=1}^D \Lc^{\text{dis}}\left( f^{\text{tea}}(\xv_k), f^{\text{stu}}(\xv_k;\tilde{\thetav}) \right) \right] \\ 
    & + \lambda_1 \cdot \left( s(\alphav)-t \right) + \lambda_2 \cdot \left( s(\alphav) - t \right)^2,
\end{split}
\label{eq:minimax}
\end{align}
where $\lambda_1, \lambda_2 \in \Rb$ are Lagrange multipliers. This additional term penalizes the distillation loss and forces the student model to meet our target sparsity. Eq.~\eqref{eq:minimax} is our training objective for Step 1 (Figure~\ref{fig:training-step1}). For Step 2 (Figure~\ref{fig:training-step2}), the objective is simply minimizing the distillation loss in Eq.~\eqref{eq:distill-loss} without any constraint because the student architecture is already fixed.

% Table
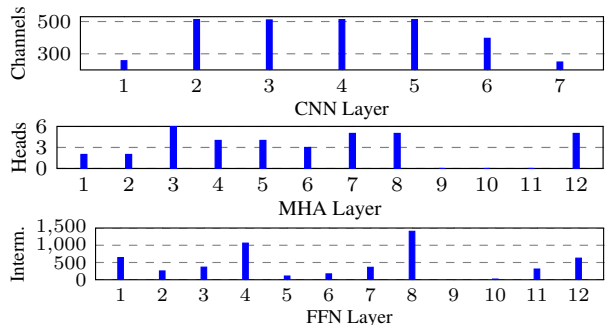
\begin{figure}[t]
     \begin{subfigure}[b]{\linewidth}
          \centering
          \resizebox{\linewidth}{!}{\begin{tikzpicture}
    \begin{axis}[
            ybar,
            enlarge x limits=true,
            ymajorgrids=true,
            grid style={dashed,gray},
		xlabel=CNN Layer,
		ylabel=Channels,
		xtick=data,
            ytick={300, 500},
            bar width=2pt,
            tickwidth=0pt,
            ymin=200,
            ymax=530,
            ylabel shift = -2pt,
            xlabel shift = -3pt,
            label style = {font=\scriptsize},
            ticklabel style={font=\scriptsize},
            width=\linewidth,
            height=0.28\linewidth,
            legend cell align={left},
            legend columns=-1,
            legend style={at={(0.3,1.05)},anchor=south west,nodes={scale=0.7, transform shape}},
            legend image code/.code={
        \draw [#1] (0pt,-1.5pt) rectangle (2pt,2pt); },
		]

        \addplot[color=blue, fill=blue] coordinates {
(1, 257)
(2, 512)
(3, 510)
(4, 512)
(5, 512)
(6, 396)
(7, 249)
	};

	\end{axis}
 
\end{tikzpicture}}  
     \end{subfigure}
     
     \vskip -0.1in
     \begin{subfigure}[b]{\linewidth}
          \centering
          \resizebox{\linewidth}{!}{\begin{tikzpicture}
    \begin{axis}[
            ybar,
            enlarge x limits={abs=9pt},
            ymajorgrids=true,
            grid style={dashed,gray},
		xlabel=MHA Layer,
		ylabel=Heads,
		xtick=data,
            ytick={0,3,6},
            label style = {font=\scriptsize},
            ticklabel style={font=\scriptsize},
            bar width=2pt,
            tickwidth=0pt,
            ymin=0,
            ymax=6,
            ylabel shift = -3pt,
            xlabel shift = -3pt,
            width=\linewidth,
            height=0.26\linewidth,
            legend cell align={left},
            legend columns=-1,
            legend style={at={(0,1)},anchor=south west,nodes={scale=0.7, transform shape}},
            legend image code/.code={
        \draw [#1] (0pt,-1.5pt) rectangle (2pt,2pt); },
		]

        \addplot[color=blue, fill=blue] coordinates {
(1,  2)
(2,  2)
(3,  6)
(4,  4)
(5,  4)
(6,  3)
(7,  5)
(8,  5) 
(9,  0)
(10, 0)
(11, 0)
(12, 5)
	};

        \legend{};
        
	\end{axis}
 
\end{tikzpicture}}  
     \end{subfigure}
     
     \vskip -0.1in
     \begin{subfigure}[b]{\linewidth}
          \centering
          \resizebox{\linewidth}{!}{\begin{tikzpicture}
    \begin{axis}[
            ybar,
            enlarge x limits={abs=9pt},
            ymajorgrids=true,
            grid style={dashed,gray},
		xlabel=FFN Layer,
		ylabel=Interm.,
		xtick=data,
            ytick={0,500,1000,1500},
            label style = {font=\scriptsize},
            ticklabel style={font=\scriptsize},
            bar width=2pt,
            tickwidth=0pt,
            ymin=0,
            ymax=1500,
            ylabel shift = -2pt,
            xlabel shift = -3pt,
            width=\linewidth,
            height=0.28\linewidth,
            legend cell align={left},
            legend columns=-1,
            legend style={at={(0,1)},anchor=south west,nodes={scale=0.7, transform shape}},
            legend image code/.code={
        \draw [#1] (0pt,-1.5pt) rectangle (2pt,2pt); },
		]

        \addplot[color=blue, fill=blue] coordinates {
(1,  643)
(2,  255)
(3,  365)
(4,  1062)
(5,  107)
(6,  170)
(7,  361)
(8,  1405)
(9,  0)
(10, 19)
(11, 310)
(12, 626)
	};

	\end{axis}
 
\end{tikzpicture}}  
     \end{subfigure}
     \vskip -0.16in
     \caption{CNN channels, attention heads and FFN intermediate sizes of DPHuBERT compressed from HuBERT Base using LibriSpeech 960h. Their original sizes are 512, 12 and 3072, respectively. The target sparsity used for pruning is 75\%.}
     \label{fig:pruned-arch}
    \vskip -0.25in
\end{figure}

\begingroup
\setlength{\tabcolsep}{4pt}

\begin{table*}[t]
  \caption{Ablation studies on DPHuBERT compressed from HuBERT Base using LibriSpeech 960h.}
  \label{tab:ablation}
  \vskip -0.14in
  \centering
  \resizebox {0.78\linewidth} {!} {
  \begin{tabular}{lccccccccccc}
    \toprule
    \multirow{2}{*}{Method} & \#Params & KS & IC & PR & ASR w/o LM & ER & QbE & SF & SID & ASV & SD \\ \cmidrule{2-12}
    & Millions & Acc$\uparrow$ & Acc$\uparrow$ & PER$\downarrow$ & WER$\downarrow$ & Acc$\uparrow$ & MTWV$\uparrow$ & F1$\uparrow$ / CER$\downarrow$ & Acc$\uparrow$ & EER$\downarrow$ & DER$\downarrow$ \\
    \midrule
    DPHuBERT & 23.59 & \textbf{96.36} & \textbf{97.92} & 9.67 & \textbf{10.47} & \textbf{63.16} & 0.0693 & \textbf{86.86} / \textbf{28.26} & \textbf{76.83} & \textbf{5.84} & \textbf{5.92} \\
    ~~w/o training step 2 & 23.59 & 94.87 & 96.76 & 10.42 & 11.55 & 62.54 &	0.0624 & 86.12 / 30.15 & 71.42 & 6.36 & 6.65\\
    ~~w/ pred-layer distill & 23.65 & 95.55 & 94.54 & 16.09 & 14.65 & 59.06 & 0.0519 & 81.83 / 36.44 & 59.88 & 7.32 & 6.75\\
    ~~w/o pruning CNN & 23.59 & 96.30 & 97.60 & \textbf{9.63} & 11.00 & \textbf{63.16} & \textbf{0.0717} & 85.77 / 29.06 & 75.22 & 6.13 & 6.10\\
    \bottomrule
  \end{tabular}
  }
  \vskip -0.1in
\end{table*}

\endgroup

\begin{figure*}[t]
     \begin{subfigure}[b]{0.33\linewidth}
          \centering
          \begin{tikzpicture}
	\begin{axis}[
		xlabel=Params (M),
		ylabel=\% Acc ($\uparrow$),
		xtick={10,30,...,100},
            ytick={94,96,...,100},
    	xmin=5,
            xmax=100,
            ymin=93,
            ymax=99,
            ylabel shift = -3pt,
            xlabel shift = -4pt,
            label style={font=\footnotesize},
            ticklabel style={font=\scriptsize},
            width=\linewidth,
            height=0.55\linewidth,
	    every axis plot/.append style={thick},
            legend cell align={left},
            legend columns=1,
            legend style={at={(1,0)},anchor=south east,nodes={scale=0.65, transform shape}}
		]

        \addplot[color=black, dotted, mark=triangle*, mark options={scale=1.5, solid, fill=black}] coordinates {
(-100, 98.34)
(94.70, 98.34)
(1000, 98.34)
	};
        \addlegendentry{HuBERT Base};

        \addplot[color=blue, solid, mark=star, mark options={scale=1, solid}] coordinates {
(23.49, 94.99)
	}; 
        \addlegendentry{DistilHuBERT};
        
	\addplot[color=green!60!black, solid, mark=*, mark options={scale=0.7, solid}] coordinates {
(89.66, 98.34)
(84.94, 98.21)
(80.23, 98.08)
(66.05, 97.76)
(47.28, 97.84)
(23.59, 97.92)
(18.87, 97.10)
(14.14, 97.18)
(9.41, 93.59)
	}; 
        \addlegendentry{DPHuBERT (ours)};

	\end{axis}
\end{tikzpicture}
          \vskip -0.13in
          \caption{IC}
     \end{subfigure}
     \begin{subfigure}[b]{0.33\linewidth}
          \centering
          \begin{tikzpicture}
	\begin{axis}[
		xlabel=Params (M),
		ylabel=\% WER ($\downarrow$),
		xtick={10,30,...,100},
            ytick={5,10,...,20},
    	xmin=5,
            xmax=100,
            ymin=5,
            ymax=17,
            ylabel shift = -3pt,
            xlabel shift = -4pt,
            label style={font=\footnotesize},
            ticklabel style={font=\scriptsize},
            width=\linewidth,
            height=0.55\linewidth,
	    every axis plot/.append style={thick},
            legend cell align={left},
            legend columns=1,
            legend style={at={(1,1)},anchor=north east,nodes={scale=0.6, transform shape}}
		]

        \addplot[color=black, dotted, mark=triangle*, mark options={scale=1.5, solid, fill=black}] coordinates {
(-100, 6.42)
(94.70, 6.42)
(1000, 6.42)
	};
        \addlegendentry{HuBERT Base};

        \addplot[color=blue, solid, mark=star, mark options={scale=1, solid}] coordinates {
(23.49, 13.37)
	}; 
        \addlegendentry{DistilHuBERT};
        
	\addplot[color=green!60!black, solid, mark=*, mark options={scale=0.7, solid}] coordinates {
(89.66, 6.55)
(84.94, 6.80)
(80.23, 7.36)
(66.05, 8.61)
(47.28, 9.25)
(23.59, 10.47)
(18.87, 11.34)
(14.14, 13.08)
(9.41, 16.20)
	}; 
        \addlegendentry{DPHuBERT (ours)};

	\end{axis}
\end{tikzpicture}
          \vskip -0.13in
          \caption{ASR}
     \end{subfigure}
     \begin{subfigure}[b]{0.33\linewidth}
          \centering
          \begin{tikzpicture}
	\begin{axis}[
		xlabel=Params (M),
		ylabel=\% Acc ($\uparrow$),
		xtick={10,30,...,100},
            ytick={50, 60, ..., 100},
    	xmin=5,
            xmax=100,
            ymin=55,
            ymax=85,
            ylabel shift = -3pt,
            xlabel shift = -4pt,
            label style={font=\footnotesize},
            ticklabel style={font=\scriptsize},
            width=\linewidth,
            height=0.55\linewidth,
	    every axis plot/.append style={thick},
            legend cell align={left},
            legend columns=1,
            legend style={at={(1,0)},anchor=south east,nodes={scale=0.65, transform shape}}
		]

        \addplot[color=black, dotted, mark=triangle*, mark options={scale=1.5, solid, fill=black}] coordinates {
(-100, 81.42)
(94.70, 81.42)
(1000, 81.42)
	};
        \addlegendentry{HuBERT Base};

        \addplot[color=blue, solid, mark=star, mark options={scale=1, solid}] coordinates {
(23.49, 73.54)
	}; 
        \addlegendentry{DistilHuBERT};
        
	\addplot[color=green!60!black, solid, mark=*, mark options={scale=0.7, solid}] coordinates {
(89.66, 81.63)
(84.94, 82.63)
(80.23, 81.74)
(66.05, 81.82)
(47.28, 81.97)
(23.59, 76.83)
(18.87, 73.07)
(14.14, 67.68)
(9.41, 59.10)
	}; 
        \addlegendentry{DPHuBERT (ours)};

	\end{axis}
\end{tikzpicture}
          \vskip -0.13in
          \caption{SID}
     \end{subfigure}
     \vskip -0.15in
     \caption{DPHuBERT trained with different target sparsities. Models are compressed from HuBERT Base using LibriSpeech 960h.}
     \label{fig:sparsities}
    \vskip -0.125in
\end{figure*}
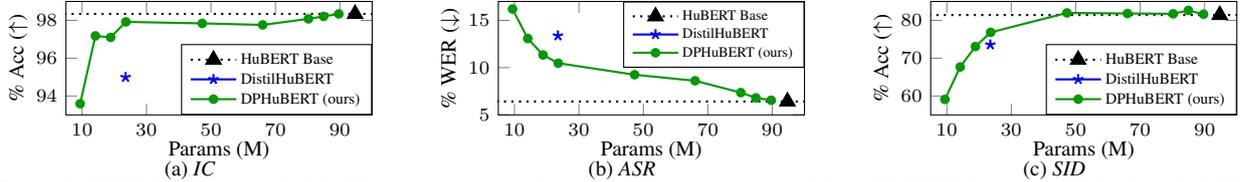

\begingroup
\setlength{\tabcolsep}{5pt}

\begin{table*}[t!]
  \caption{Compressing HuBERT Large to have a similar size as HuBERT Base using LibriSpeech 960h.}
  \label{tab:hubert-large}
  \vskip -0.14in
  \centering
  \resizebox {0.78\linewidth} {!} {
  \begin{tabular}{lccccccccccc}
    \toprule
    \multirow{2}{*}{Method} & \#Params & KS & IC & PR & ASR w/o LM & ER & QbE & SF & SID & ASV & SD \\ \cmidrule{2-12}
    & Millions & Acc$\uparrow$ & Acc$\uparrow$ & PER$\downarrow$ & WER$\downarrow$ & Acc$\uparrow$ & MTWV$\uparrow$ & F1$\uparrow$ / CER$\downarrow$ & Acc$\uparrow$ & EER$\downarrow$ & DER$\downarrow$ \\
    \midrule
    HuBERT Large & 316.60 & 95.29 & 98.76 & 3.53 & 3.62 & 67.62 & 0.0353 &	89.81 / 21.76 & 90.33 & 5.98 & 5.75\\
    HuBERT Base & 94.68 & 96.30 & 98.34 & 5.41 & 6.42 & 64.92 & 0.0736 & 88.53 / 25.20 & 81.42 & 5.11 & 5.88 \\
    \midrule
    DPHuBERT & 94.59 & 94.51 & 98.47 & 4.46 & 6.23 & 65.11 & 0.0246 & 88.37 / 24.60 & 83.17 & 7.05 & 5.79\\
    \bottomrule
  \end{tabular}
  }
  \vskip -0.25in
\end{table*}

\endgroup

\section{Experiments}

\subsection{Experimental setup}

\noindent\textbf{Toolkits.}
Our method is implemented with PyTorch~\cite{pytorch} and TorchAudio~\cite{torchaudio}. Pre-trained SSL models are downloaded from fairseq~\cite{fairseq} or Hugging Face~\cite{huggingface}.

\noindent\textbf{Data.}
The unlabeled LibriSpeech 960h~\cite{librispeech-corpus} is used by default. In Section~\ref{subsec:main-results} and Table~\ref{tab:main-results}, the train-clean 100h subset is also used to investigate the effect of training data size.

\noindent\textbf{Model.}
In the default setup, we compress HuBERT Base~\cite{hubert}. To verify the generalizability, we also compress WavLM Base+~\cite{wavlm} in Section~\ref{subsec:main-results} and HuBERT Large~\cite{hubert} in Section~\ref{subsec:compress-large}.

\noindent\textbf{Training.}
DPHuBERT is trained on 4 NVIDIA A100 (40GB) GPUs with 640 seconds of audio per mini-batch. In Step 1, the peak learning rates of main parameters $\thetav$ and auxiliary parameters $\alphav, \lambdav$ are 2e-4 and 2e-2, respectively. Warmup and total steps are 15k and 50k, respectively. The target sparsity $t$ is linearly increased to the desired value in 5k steps, which facilitates training~\cite{yifan-pruning}. In Step 2, the peak learning rate is 1e-4. Warmup and total steps are 5k and 25k, respectively. In the default setup, the total training time of DPHuBERT is only 6 hours, i.e., 24 GPU hours (due to our usage of 4 GPUs).

\noindent\textbf{Evaluation.}
The SUPERB~\cite{superb} benchmark consists of 10 tasks: keyword spotting (KS), intent classification (IC), phoneme recognition (PR), ASR, emotion recognition (ER), query by example (QbE), slot filling (SF), speaker identification (SID), automatic speaker verification (ASV) and speaker diarization (SD). We follow their default configurations in all tasks except that SID uses a learning rate of 5e-3.

\subsection{Main results}
\label{subsec:main-results}

Table~\ref{tab:main-results} compares various methods on SUPERB~\cite{superb}. DPHuBERT is compressed from HuBERT Base. With LibriSpeech 960h, DPHuBERT outperforms pure distillation methods (including DistilHuBERT~\cite{distilhubert}, FitHuBERT~\cite{fithubert}, FitW2V2~\cite{fithubert} and two best-performing models from~\cite{deep-vs-wide}) in 8 out of 10 tasks. This shows that DPHuBERT better preserves the general speech representations of the teacher model, covering content, speaker and semantics. With only 100h training data, DPHuBERT still performs much better than prior methods in almost all tasks. Surprisingly, DPHuBERT using 100h already outperforms previous distilled models using 960h in IC, PR, QbE and SD, and has similar results in the other tasks. This shows that DPHuBERT learns powerful representations even from limited data.

We have also compressed WavLM Base+ to obtain DPWavLM. Compared to DPHuBERT, DPWavLM achieves further improvements in 8 tasks. This is because the unpruned WavLM Base+ is better than the unpruned HuBERT Base. These results demonstrate that our compression method can be applied to different speech SSL models.

Figure~\ref{fig:pruned-arch} shows the architecture of DPHuBERT, which is automatically discovered by structured pruning. For CNN, the first and last layers are pruned the most. For MHA, three higher layers are entirely removed, indicating those layers are more redundant. Our results are consistent with prior studies about pruning~\cite{cofi, yifan-pruning} or general speech encoders~\cite{diagnality, shim2022understanding, branchformer, att-w-smoothing}. For FFN, the 4th, 8th and 12th layers are preserved more than their neighbors, because those layers are explicitly matched between the teacher and student models as defined by Eq.~\eqref{eq:distill-loss} in Section~\ref{subsec:distill-loss}.

\subsection{Ablation studies}
\label{subsec:ablation-studies}

Table~\ref{tab:ablation} summarizes the results of the following ablation studies.

\noindent\textbf{Two-step training.}
DPHuBERT has two training steps (Section~\ref{subsec:training-procedure} and Figure~\ref{fig:training-steps}). The pruned model after Step 1 without Step 2 is evaluated in the second row of Table~\ref{tab:ablation}. It is worse than DPHuBERT in all tasks, verifying the necessity of Step 2. This is because Step 1 optimizes Eq.~\eqref{eq:minimax} where the regularization term competes with the distillation loss to meet the target sparsity, while Step 2 directly optimizes the distillation loss to improve the student's learned representations.

\noindent\textbf{Distillation methods.}
As discussed in Section~\ref{subsec:background-distill}, DPHuBERT uses layer-to-layer distillation instead of prediction-layer distillation in DistilHuBERT~\cite{distilhubert}. The third row of Table~\ref{tab:ablation} shows that prediction-layer distillation causes severe degradations in all tasks, probably due to the deep student architecture. Directly matching intermediate layers facilitates the training of deep students as found in~\cite{deep-vs-wide}.

\noindent\textbf{Pruning units.}
DPHuBERT prunes both CNN and Transformer because CNN has a high computational cost~\cite{yifan-pruning,melhubert}. The fourth row of Table~\ref{tab:ablation} shows results without pruning CNN (i.e., only pruning attention heads and FFN intermediate sizes). This model is (slightly) worse than the default setup in 7/10 tasks. This verifies that the CNN also has redundant components which can be pruned, as reported in~\cite{wu2022performance, yifan-pruning, melhubert}.

\subsection{Results at various sparsities}
We train DPHuBERT with various target sparsities ($t$ in Eqs.~\eqref{eq:eq-constraint}~\eqref{eq:minimax}) and show results in Figure~\ref{fig:sparsities}. For IC and SID, our method can significantly reduce the model size while keeping a similar accuracy as the original HuBERT Base. For ASR, the degradation is more severe, probably because the sequence transduction task is more challenging than classification tasks.

\subsection{Compressing HuBERT Large}
\label{subsec:compress-large}

Our method can be applied to larger speech SSL models with very limited training cost. In Table~\ref{tab:hubert-large}, HuBERT Large is compressed to have a similar size as HuBERT Base, which only takes about 60 GPU hours. The compressed model even outperforms HuBERT Base in several tasks like PR, SF-CER and SID. It is worse than HuBERT Base in KS, QbE and ASV, but the teacher model, HuBERT Large, is also clearly worse than HuBERT Base in those tasks.

\section{Conclusion}

This work proposes DPHuBERT, a task-agnostic compression method based on joint distillation and structured pruning. DPHuBERT outperforms previous distillation methods in most tasks of SUPERB. Comprehensive analyses are presented to investigate its performance with less training data or at various sparsity ratios. In addition to HuBERT Base, our method can be directly applied to other speech SSL models such as WavLM and HuBERT Large while still being efficient and effective.
In the future, we will exlore more sophisticated distillation objectives (e.g., the masking-based distillation loss used in LightHuBERT~\cite{lighthubert}) to further improve the performance.

\section{Acknowledgements}
We use PSC Bridges2 and NCSA Delta via ACCESS allocation CIS210014, supported by National Science Foundation grants \#2138259, \#2138286, \#2138307, \#2137603, and \#2138296.

\bibliographystyle{IEEEtran}
\bibliography{mybib}

\end{document}